\documentclass[12pt, a4paper]{elsarticle}
\usepackage{graphicx}%
\usepackage{multirow}%
\usepackage{amsmath, amssymb,amsfonts}%
\usepackage{amsthm}%
\usepackage{mathrsfs}%
\usepackage[title]{appendix}%
\usepackage{xcolor}%
\usepackage{textcomp}%
\usepackage{manyfoot}%
\usepackage{booktabs}%
\usepackage{algorithm}%
\usepackage{algorithmicx}%
\usepackage{algpseudocode}%
\usepackage{listings}%
\usepackage{url}
\usepackage{pdfpages}

\usepackage[all]{xy}
\usepackage{calc}
\usepackage{tikz}
\usetikzlibrary{arrows.meta, positioning}

\journal{$$}

\usepackage{amsthm}%
\newtheorem{lemma}{Lemma}
\newtheorem{theorem}{Theorem}
\newtheorem{definition}{Definition}

\newcommand{\R}{\mathbb{R}}

\begin{document}

\begin{frontmatter}

\title{Fully tensorial approach to hypercomplex-valued neural networks}
\author[inst1]{Agnieszka Niemczynowicz\corref{cor1}}
\cortext[cor1]{Corresponding author}
\ead{aga.niemczynowicz@gmail.com}

\author[inst1]{Rados\l{}aw Antoni Kycia}
\ead{kycia.radoslaw@gmail.com}

\affiliation[inst1]{organization={Faculty of Computer Science and Mathematics, Cracow University of Technology},
            addressline={Warszawska 24}, 
            city={Cracow},
            postcode={PL31-155}, 
            country={Poland}}

\begin{abstract}
We present a fully tensorial theoretical framework for hypercomplex-valued neural networks.
The proposed approach enables neural network architectures to operate on data defined
within arbitrary finite-dimensional algebras. The central observation is that algebra
multiplication can be represented by a rank-three tensor, which allows all algebraic
operations in neural network layers to be formulated in terms of standard tensor
contractions, permutations, and reshaping operations.

This tensor-based formulation provides a unified and dimension-independent description of hypercomplex-valued dense and convolutional layers and is directly compatible with modern deep learning libraries that support optimized tensor operations. The proposed framework
recovers existing constructions for four-dimensional algebras as special cases.

Within this setting, we establish a tensor-based version of the universal approximation theorem for single-layer hypercomplex-valued perceptrons under mild non-degeneracy assumptions on the underlying algebra, thereby providing a rigorous theoretical foundation for the
considered class of neural networks.
\end{abstract}

\end{frontmatter}



\section{Introduction}\label{sec1}

The rapid development of Artificial Neural Networks (NNs) has stimulated new research directions and theoretical generalizations. These developments increasingly involve advanced mathematical concepts, including group theory \cite{GroupTheoryNN}, differential geometry \cite{GeometricDeepLeanring, GeoemtricDeepLearning2}, and topological methods in data analysis \cite{ToplogicalDataAnalysis}.

The core of NN implementations relies on linear algebra. In widely used programming libraries, such as \textit{TensorFlow} \cite{Tensorflow} and \textit{PyTorch} \cite{PyTorch}, feed-forward neural networks are commonly implemented as compositions of layers through which data are propagated in a prescribed direction. This propagation is realized by optimized tensor operations on multidimensional arrays. Modern deep learning libraries provide substantial computational acceleration for multidimensional tensor operations \cite{tenosorialGPU, tensorialGPU2}. Consequently, reformulating neural network computations in compact tensorial notation is not only mathematically natural, but also relevant from the computational point of view.

Several algebraic extensions of neural networks have been proposed. One direction is represented by Algebraic Neural Networks \cite{AlgebraicNeuralNetworks}, where additional endomorphism operations are applied to data. Another algebraic-geometric direction concerns neural networks based on Geometric Algebra or Clifford algebra \cite{CliffordNN, GeometricCliffordNeuralNetworks}, with quaternion-valued neural networks as one of the most prominent examples \cite{QuaternionicNN2, QuaternionicNN1}. Recently, Parametrized Hypercomplex-valued Neural Networks have been introduced \cite{ParametrizedHypercomplexNN, PHN_Zhang} for convolutional layers. These models can learn an optimal hypercomplex algebra adapted to the data by exploiting an optimized Kronecker-product structure. However, in some applications, the hypercomplex algebra, or more general algebraic parameters, are treated as fixed hyperparameters. This makes it possible to study and optimize the algebraic structure at the meta-level.

In this paper, we develop a formulation in which classical real-valued algebraic computations are replaced by hypercomplex-valued, or more general algebra-valued, computations by representing algebra multiplication as a tensorial operation. The hypercomplex-valued NN approach is not new, e.g., \cite{BookHypercomplexNN, QuaternionicNN2}. However, interest in this direction has recently increased due to favorable trainable-parameter complexity properties reported in areas such as image processing \cite{Marcos1} and time series analysis \cite{NiemczynowiczTimeSeries}. In these contributions, open-source implementations \cite{Marcos1} for specific four-dimensional hypercomplex algebras were used. Hypercomplex-valued algorithms have also been applied in nonlinear adaptive filtering \cite{nonlinearAdaptiveFiltering}, recommendation systems \cite{recommendation, recommendationSystem}, graph classification \cite{GraphsLearning}, drone control \cite{HypercomplexRobotics}, knowledge graph embedding \cite{KnowledgeGraphsEmbedding1}--\cite{KnowledgeEmbeddingGraphs3}, knowledge graph completion \cite{KnowledgeGraphCompletion}, multi-view medical image classification \cite{BreastCancer}, speech recognition \cite{speechRecognition}, and image inpainting \cite{imageInpainting}. Related architectures include hypercomplex recurrent convolutional architectures \cite{HypercomplexRCNN} and quaternionic convolutional neural networks \cite{QCNN}. These architectures are typically restricted to specific algebras adjusted to the dimensionality and structure of the data. Existing results suggest that hypercomplex-valued architectures may provide improved performance, or comparable performance with fewer trainable parameters, relative to analogous real-valued architectures. The development of optimized and scalable approaches to hypercomplex-valued neural networks therefore remains an important research direction.

The implementation presented in this article significantly extends the ideas from \cite{Marcos1, QuaternionicNN2} to arbitrary algebras, including hypercomplex ones. The algorithms described here agree with the NNs presented in \cite{Marcos1, QuaternionicNN2} when restricted to four-dimensional hypercomplex algebras. Four-dimensional hypercomplex algebras are useful, for example, for encoding RGB color data in images. Moreover, the implementation in \cite{Marcos1} relies on constructing an additional multiplication matrix from the multiplication table of the hypercomplex algebra. This introduces an additional step in setting up a neural network. In contrast, the present approach avoids this construction and generalizes the formulation to arbitrary finite-dimensional algebras, since the multiplication table of the algebra is used directly, as given in standard references, e.g., \cite{AlgebraChapter0}. Moreover, the restriction to four-dimensional algebras can be relaxed. This is important for processing data that are naturally represented as $n$-tuples ($n>0$, integer), which may be encoded as single elements of an $n$-dimensional algebra. This constitutes one of the main theoretical contributions of the present work to the general algebraic formulation of hypercomplex-valued neural networks.

\par The present work is intentionally focused on the theoretical and algorithmic foundations of hypercomplex-valued neural networks formulated in a fully tensorial framework. Practical implementation details, software architecture, reference implementations in TensorFlow and PyTorch, as well as illustrative experimental studies validating the proposed dense and convolutional layers, have been reported separately in \cite{NiemczynowiczKycia}. That contribution provides executable examples and empirical training pipelines confirming the correctness and usability of the tensor-based formalism introduced here. By separating the theoretical framework from implementation-oriented validation, the present paper aims to establish a general, algebra-agnostic foundation that can be reused independently of specific software realizations.

The main contributions of this paper are as follows: 
\begin{itemize}
    \item the establishment of a general tensorial representation of algebra multiplication as a third-rank tensor, providing a unified mathematical foundation for hypercomplex-valued neural network operations,
    \item the formulation of a general algorithm for computations within the hypercomplex-valued dense layer that generalizes the ideas of \cite{Marcos1, QuaternionicNN2} to arbitrary finite-dimensional algebras,
    \item the formulation of general algorithms for 1-, 2-, and 3-dimensional hypercomplex-valued convolutional layer computations that generalize the ideas of \cite{Marcos1, QuaternionicNN2} to arbitrary finite-dimensional algebras. 
\end{itemize}

The extension of neural network layers to arbitrary $n$-dimensional algebras ($n>0$) allows $n$-tuples of data to be represented as single entities processed by NN layers, which is the central idea of hypercomplex-valued NNs. For example, a color pixel consisting of three numbers encoding RGB components can be treated as a single element of a four-dimensional algebra padded with an alpha channel or a null value, rather than as three separate numbers in a real-valued NN. Numerical experiments reported in the literature indicate that hypercomplex-valued NNs may require fewer trainable parameters to achieve comparable accuracy to real-valued NNs, see e.g., \cite{Marcos1, NiemczynowiczTimeSeries}.

The paper is organized as follows. After recalling the general algebraic notions 
in the next section, we present the algorithms for general architectures of dense and convolutional hypercomplex-valued NNs. Finally, in Section~\ref{Concl}, conclusions and future research directions are discussed. Supplementary algebraic background and computational remarks are collected in
Appendices~\ref{app:algebra_tensor} and~\ref{app:computational}, respectively.

\section{Theory}\label{sec2}
This section provides an overview of the mathematical theory underlying the operations used to implement hypercomplex-valued neural networks. These notions constitute the methodological foundation of the present paper. A tensor is a classical object of multilinear algebra, discussed in detail in standard references, e.g., \cite{AlgebraChapter0}. We recall the relevant notions here in order to prepare the background for expressing algebra multiplication as a third-rank tensor.

\subsection{Tensors}

The primary object used in modern neural network implementations is a \emph{tensor}, which in machine learning libraries such as \textit{TensorFlow} \cite{Tensorflow} and \textit{PyTorch} \cite{PyTorch} denotes a multidimensional array of numerical values. In mathematics, tensors are multilinear objects defined independently of a particular coordinate representation and acquire an array form only after fixing bases in the underlying vector spaces. In this work, we explicitly distinguish between these two notions.

Let $V$ be a finite-dimensional vector space over a field $\mathbb{F}$ and $V^{*}$ its dual. Tensor products of vector spaces are used to represent multilinear mappings in a coordinate-free manner \cite{AlgebraChapter0, PenroseSpinors}. For vector spaces $V_{1},\ldots,V_{k}$, their tensor product is denoted by $V_{1}\otimes\cdots\otimes V_{k}$. Once bases are fixed, tensors can be represented by collections of components arranged as multidimensional arrays.

In abstract index notation, a $(p,q)$-tensor is written as \cite{AlgebraChapter0}
\begin{equation}
T = T^{j_{1}\ldots j_{q}}_{i_{1}\ldots i_{p}}\,
e^{i_{1}}\otimes\cdots\otimes e^{i_{p}}\otimes
e_{j_{1}}\otimes\cdots\otimes e_{j_{q}},
\end{equation}
where upper and lower indices indicate contravariant and covariant components, respectively. In numerical libraries, these components are stored as multidimensional tensors, while the transformation properties associated with index positions are not explicitly enforced.

In practical implementations, it is convenient to describe tensors by their \emph{shape}, understood as an ordered tuple of index ranges. Each axis of a numerical tensor corresponds to a factor in the underlying tensor product.

\vspace{2mm}

\noindent\textbf{Permutation.}
For a permutation $p$ of $\{1,\ldots,k\}$, the permutation operator $\tau_{p}$ reorders tensor factors as \cite{AlgebraChapter0}
\begin{equation}
\tau_{p}(v_{1}\otimes\cdots\otimes v_{k})
=
v_{p(1)}\otimes\cdots\otimes v_{p(k)}.
\end{equation}
In abstract index notation, this corresponds to a reordering of indices.

\vspace{2mm}

\noindent\textbf{Reshaping.}
Reshaping combines multiple indices into a single index without changing the underlying data. For indices $i$ and $j$, we define \cite{AlgebraChapter0}
\begin{equation}
Rs(i,j)= i\,R(j)+j,
\end{equation}
where $R(j)$ denotes the range of index $j$. This operation modifies only the indexing structure and is used to interface tensor contractions with matrix-based computations.

\vspace{2mm}

\noindent\textbf{Contraction.}
Given tensors with compatible covariant and contravariant indices, contraction corresponds to summation over a matched index pair. In abstract index notation \cite{AlgebraChapter0}, 
\begin{equation}
C_{p,q}(A,B)
=
A_{\ldots i_{p} \ldots}\, B^{\ldots i_{q} \ldots},
\end{equation}
where implicit summation over the repeated index is assumed. Contraction is the key operation used to express algebra multiplication in tensor form.

\vspace{2mm}

\noindent\textbf{Concatenation.}
Concatenation joins tensors of identical shape along a specified axis, forming a larger tensor whose corresponding dimension equals the sum of the original dimensions.

Detailed definitions of the tensor product, its universal factorization property, and additional examples are provided in Appendix~A.

\subsection{(Hypercomplex) Algebras}
\label{Subsection_HypercomplexAlgebra}

In this section, we introduce the algebraic structures underlying hypercomplex-valued neural networks and formulate algebra multiplication in a tensorial form suitable for neural network implementations.

\vspace{2mm}

\noindent\textbf{Definition.}
An algebra over a field $\mathbb{F}$ is a vector space $V$ equipped with a bilinear product
$\cdot : V \times V \rightarrow V$
satisfying
\begin{itemize}
 \item[a)] $(x+y)\cdot z = x\cdot z + y\cdot z,$
 \item[b)] $z\cdot (x+y) = z\cdot x + z\cdot y,$
 \item[c)] $(\alpha x)\cdot(\beta y) = \alpha\beta (x\cdot y),$
\end{itemize}
for all $x,y,z \in V$ and $\alpha,\beta \in \mathbb{F}$.
The algebra is called \emph{commutative} if $x\cdot y = y\cdot x$ for all $x,y \in V$.

Throughout the paper, we assume that the algebra admits a neutral element, denoted by $1$, and that a basis $\{e_{i}\}_{i=0}^{n}$ of $V$ is chosen such that $e_{0}=1$.

\vspace{2mm}

Efficient use of algebras in tensor-based numerical frameworks such as \textit{TensorFlow} and \textit{PyTorch} relies on representing algebra multiplication as a multilinear operation. This leads to the following fundamental result.

\vspace{2mm}

\noindent\textbf{Theorem.}
Let $V$ be a finite-dimensional algebra over $\mathbb{F}$. The multiplication in $V$ can be represented by a third-rank tensor
\[
A \in V^{*}\otimes V^{*}\otimes V.
\]
For a fixed basis $V=\mathrm{span}\{e_{i}\}_{i=0}^{n}$ and the dual basis $\{e^{i}\}$ of $V^{*}$, the tensor has the form
\begin{equation}
A = A^{k}_{ij}\, e^{i}\otimes e^{j}\otimes e_{k}.
\end{equation}

The product of elements $x = x^{i}e_{i}$ and $y = y^{j}e_{j}$ in $V$ is then given by
\begin{equation}
x\cdot y := A(x,y) = A^{k}_{ij} x^{i} y^{j} e_{k}.
\end{equation}

\begin{proof}
The result follows directly from the bilinearity of the algebra multiplication, which guarantees the existence of a unique tensor representation in $V^{*}\otimes V^{*}\otimes V$.
\end{proof}

\vspace{2mm}

\noindent\textbf{Corollary.}
If the algebra is commutative, then the multiplication tensor is symmetric with respect to its covariant indices, i.e.,
\[
A^{k}_{ij} = A^{k}_{ji}.
\]

\vspace{2mm}

The coefficients $A^{k}_{ij}$ coincide with the structure constants of the algebra in the chosen basis \cite{FultonHarris}. In numerical implementations, these coefficients are stored as multidimensional arrays (referred to as \emph{tensors} in machine learning libraries), forming the core object used to implement hypercomplex-valued neural network layers.

\vspace{2mm}

\noindent\textbf{Example.}
For the field of complex numbers $\mathbb{C}$, viewed as a two-dimensional real algebra with basis $\{e_{0},e_{1}\}$, the multiplication is given by
\[
e_{0}\cdot e_{0}=e_{0}, \quad
e_{0}\cdot e_{1}=e_{1}, \quad
e_{1}\cdot e_{0}=e_{1}, \quad
e_{1}\cdot e_{1}=-e_{0}.
\]
The corresponding multiplication tensor is
\begin{equation}
A_{\mathbb{C}} =
e^{0}\otimes e^{0}\otimes e_{0}
+ e^{0}\otimes e^{1}\otimes e_{1}
+ e^{1}\otimes e^{0}\otimes e_{1}
- e^{1}\otimes e^{1}\otimes e_{0}.
\end{equation}

\section{Neural Network Architectures}
\label{sec3}

In this section, we present the mathematical formulation of hypercomplex-valued dense and convolutional neural network layers.

Throughout this section, we do not distinguish between covariant and contravariant indices and write all indices in subscript form. Moreover, tensors are treated as multidimensional arrays with indices starting from zero. This convention is consistent with standard implementations in tensor-based libraries such as \textit{TensorFlow} and \textit{PyTorch} \cite{Tensorflow, PyTorch}.

We assume that the algebra multiplication structure constants are fixed and stored in a third-rank tensor
$A = A_{i_{a} j_{a} k_{a}}$
(in abstract index notation), as introduced in Subsection~\ref{Subsection_HypercomplexAlgebra}.

The architectures and algorithms presented below constitute a generalization of hypercomplex-valued neural network constructions originally developed for four-dimensional algebras \cite{Marcos1,QuaternionicNN2}. In contrast to fixed-dimensional formulations, the proposed approach extends these architectures to algebras of arbitrary finite dimension by expressing algebra multiplication entirely in terms of tensor operations. These operations are directly compatible with and optimized for modern tensor computation frameworks such as \textit{TensorFlow} and \textit{PyTorch} \cite{Tensorflow, PyTorch}.

The resulting algorithms have been implemented in the \textit{Hypercomplex Keras} Python library~\cite{HypercomplexKeras} and further described in~\cite{NiemczynowiczKycia}. For four-dimensional algebras, the implementation reproduces results consistent with those reported in~\cite{Marcos1}.

\begin{figure}[htb]
\centering
\resizebox{\linewidth}{!}{%
\begin{tikzpicture}[
    node distance=2.0 cm,
    every node/.style={draw, rounded corners, align=center, minimum height=1.4cm, inner sep=4pt, font=\small},
    arrow/.style={->, thick}
]
\node (alg) {Algebra $V$\\ basis $\{e_i\}$};
\node (tensor) [right=of alg] {Multiplication tensor\\ $A_{ij}^{k}$};
\node (ops) [right=of tensor] {Tensor operations\\ contraction, permutation, reshape};
\node (layer) [right=of ops] {Hypercomplex layer\\ HyperDense / HyperConv};
\draw[arrow] (alg) -- (tensor);
\draw[arrow] (tensor) -- (ops);
\draw[arrow] (ops) -- (layer);
\end{tikzpicture}%
}
\caption{Pipeline from algebraic multiplication to hypercomplex-valued neural network layers.}
\label{fig:algebra_to_layer_1line}
\end{figure}

\newpage

\subsection{Hypercomplex-valued dense layer}
\label{HyperDense}
We begin by describing the hypercomplex-valued dense layer. It is a general-purpose neural network layer that operates on data augmented with an additional axis corresponding to algebra components. We assume that the input data have dimensions
\[
b \times \underbrace{a \times in}_{n},
\]
where $b$ denotes the batch size, $a=\dim(V)$ is the algebra dimension, and $in$ is a positive integer multiplier representing number of weights. The last two quantities determine the effective input dimension $n=a\cdot in$.

The input tensor is denoted in abstract index notation by
\[
X = X_{i_{b}\,Rs(i_{a},i_{in})},
\]
where $i_{b}$ is the batch index, $i_{a}$ indexes the algebra components, and $i_{in}$ indexes the multiplicity of the algebra dimension. The learnable parameters (kernel weights) are given by
\[
K = K_{i_{a} i_{in} i_{u}},
\]
where $i_{u}$ indexes the output units (neurons). An optional bias term
\[
b = b_{Rs(i_{a}, i_{u})}
\]
can be included. Kernel weights and biases are typically initialized using standard initialization schemes, such as those proposed in~\cite{Glorot}.

The algorithm defining the hypercomplex-valued dense layer is summarized in Algorithm~\ref{Algorithm_hyperdens}.

\begin{algorithm}
\caption{Hypercomplex-valued dense layer}
\begin{algorithmic}
\Require $X$, $A$, $\sigma$, bias, activation
\Ensure $K$, $b$ initialized
\State $W \gets C_{1,0}(A,K)$
\Statex \hspace{1em} [AIN: $W_{i_{a} k_{a} i_{in} i_{u}} \gets \sum_{j} A_{i_{a} j k_{a}} K_{j i_{in} i_{u}}$]
\State $W \gets \tau_{p=(0\!\rightarrow\!0,\;1\!\rightarrow\!2,\;2\!\rightarrow\!1,\;3\!\rightarrow\!3)}(W)$
\Statex \hspace{1em} [AIN: $W_{i_{a} i_{in} k_{a} i_{u}} \gets W_{p(i_{a} k_{a} i_{in} i_{u})}$]
\State $W \gets Rs_{1}\circ Rs_{0}(W)$
\Statex \hspace{1em} [AIN: $W_{i_{1} i_{2}} \gets W_{Rs(i_{a},i_{in})\,Rs(k_{a},i_{u})}$]
\State $Output \gets C_{1,0}(X,W)$
\Statex \hspace{1em} [AIN: $Output_{i_{b} i_{2}} \gets \sum_{k} X_{i_{b} k=Rs(i_{a},i_{in})} W_{k i_{2}}$]
\If {bias is True}
  \State $Output \gets Output + b$
\EndIf
\If {activation is True}
  \State $Output \gets \sigma(Output)$
\EndIf
\end{algorithmic}
\label{Algorithm_hyperdens}
\end{algorithm}
Figure~\ref{fig:hyperdense_indices} illustrates the sequence of index transformations
(contraction, permutation, and reshape) used to obtain the effective weight matrix
in Algorithm~\ref{Algorithm_hyperdens}.

\begin{figure}[htb]
\centering
\begin{tikzpicture}[
  node distance=1.4cm and 2.4cm,
  every node/.style={
    draw,
    rounded corners,
    align=center,
    text width=0.38\linewidth,
    minimum height=1.1cm,
    inner sep=4pt,
    font=\small
  },
  arrow/.style={->, thick}
]

\node (K) {Kernel tensor\\
$K_{\,i_{a}\, i_{in}\, i_u}$};

\node (A) [right=of K] {Algebra tensor\\
$A_{\,i_{a}\, j_{a}\, k_{a}}$};

\node (C) [below=of K] {Contraction $C_{1,0}$\\
$W_{\,i_{a}\, k_{a}\, i_{in}\, i_u}$};

\node (P) [right=of C] {Permutation $\tau$\\
$(i_{a},k_{a},i_{in},i_u)$\\
$\mapsto (i_{a},i_{in},k_{a},i_u)$};

\node (R) [below=of C] {Reshape $Rs$\\
$(i_{a},i_{in}) \mapsto j$};

\node (W) [right=of R] {Weight matrix\\
$W_{\,j\, h}$};

\draw[arrow] (K) -- (C);
\draw[arrow] (A) -- (C);

\draw[arrow] (C) -- (P);
\draw[arrow] (P) -- (R);
\draw[arrow] (R) -- (W);

\end{tikzpicture}
\caption{Index transformations in the HyperDense layer (Algorithm~\ref{Algorithm_hyperdens}.).
The algebra multiplication tensor $A_{ijk}$ is first contracted with the kernel tensor, then permuted and reshaped to obtain a standard weight matrix used in the final linear mapping.}
\label{fig:hyperdense_indices}
\end{figure}

For clarity, both tensorial notation and abstract index notation (AIN) are provided. Since all indices are written in subscript form, summation symbols are stated explicitly. Two Boolean flags are used: \texttt{bias}, indicating whether the bias term is included, and \texttt{activation}, indicating whether a nonlinear activation function $\sigma$ is applied.
 
Implementations of the hypercomplex-valued dense layer using \textit{Keras} with the \textit{TensorFlow} backend as well as \textit{PyTorch} are described in~\cite{NiemczynowiczKycia} and provided in the \textit{Hypercomplex Keras} library~\cite{HypercomplexKeras}. As emphasized above, for $4$-dimensional algebras the resulting output agrees with the formulations and results reported in~\cite{Marcos1}.

The FLOP complexity is presented in the following theorem.
\begin{theorem}
 The FLOP complexity of a hypercomplex-valued dense layer is of the following order, assuming $b> a$
\begin{equation}
    \mathcal{O}(b \cdot a^{2} \cdot u\cdot in)=\mathcal{O}(b \cdot a \cdot n \cdot u),
\end{equation} 
compared with the corresponding real-valued dense layer
\begin{equation}
    \mathcal{O}(b \cdot n \cdot u).
\end{equation}
\end{theorem}

\begin{proof}
Contraction over one index of two multidimensional arrays introduces a FLOP complexity given by the product of all non-contracted dimensions and the dimension of the contracted index.
Since the contractions introduce the most essential computational overhead, we focus on them in the Algorithm \ref{Algorithm_hyperdens}. 

The first contraction introduces the complexity $\mathcal{O}(a^{3} \cdot u \cdot in)$. The second contraction has complexity $\mathcal{O}(b \cdot a^{2} \cdot in \cdot u)$. If we assume that the batch size is larger than the algebra size $b > a$, we get the result. 
\end{proof}

For $a=1,$ the complexities for real-valued, and hypercomplex-valued dense layers coincide.
A FLOP gain is possible if the corresponding real-valued dense layer requires at least $a$ times more units $u$ than the hypercomplex-valued dense layer.

\subsection{Hypercomplex-valued convolutional layer}
\label{HyperConv}

In this subsection, we describe hypercomplex-valued convolutional neural network layers in a unified formulation for $k$-dimensional convolutions, where $k\in\{1,2,3\}$. The cases differ only by the shape of the input data and the kernel dimensions.

Additional data channels, such as RGB channels in standard image representations, can be embedded into algebra elements. For instance, a two-dimensional image can be interpreted as a matrix of elements of a four-dimensional algebra by combining color channels together with an additional alpha component, so that each pixel corresponds to a single algebra element.

The core idea of hypercomplex-valued convolution is to use the algebra multiplication tensor $A$ to separate algebra basis components and subsequently apply standard real-valued convolution independently to each resulting component. This construction generalizes the approach introduced in~\cite{Marcos1} for four-dimensional algebras and, in particular, quaternion-valued convolutional networks described in~\cite{QuaternionicNN2}. In~\cite{Marcos1}, the separation of algebra components was implemented through a fixed multiplication matrix, whereas the present formulation achieves this in a dimension-independent manner using tensor operations.

The input data tensor is denoted by
\[
X = X_{i_{b} i_{1}\ldots i_{k}\,Rs(i_{a}, i_{in})},
\]
with dimensions
\[
b \times n_{1} \times \ldots \times n_{k} \times \underbrace{a \times in}_{n},
\]
where $b$ is the batch size, $n_{i}$ ($i\in\{1,\ldots,k\}$) denote the spatial dimensions of the data sample (e.g., raster coordinates for $k=2$), $a=\dim(V)$ is the algebra dimension, and $in$ is a positive integer multiplier.\footnote{The ordering of indices follows the convention used in \textit{TensorFlow}, where spatial dimensions follow the batch dimension. In \textit{PyTorch}, this multiindex is placed at the end; adapting the formulation requires only index permutations and is therefore not discussed separately}.

The kernel has dimensions
\[
al \times L_{1} \times \ldots \times L_{k} \times in \times F,
\]
where $L_{i}$ ($i\in\{1,\ldots,k\}$) denote the kernel sizes along each axis and $F$ is the number of filters. The kernel tensor is written as
\[
K = K_{i_{a} i_{l_{1}}\ldots i_{l_{k}} i_{in} i_{f}}.
\]
An optional bias term is controlled by the flag \texttt{bias}; if used, the bias has dimension $a \times F$. Kernel weights and biases can be initialized using standard schemes, such as those proposed in~\cite{Glorot}.

We denote by $convkD(X,K,\text{strides},\text{padding})$ the standard optimized $k$-dimensional convolution operation applied component-wise to algebra elements, as implemented in modern deep learning frameworks~\cite{Convolution}. The complete algorithm is summarized in Algorithm~\ref{Algorithm_hyperKconv}.

\begin{algorithm}
\caption{Hypercomplex-valued $k$-dimensional convolutional layer}
\begin{algorithmic}
\Require $X$, $A$, $\sigma$, bias, activation
\Ensure $K$, $b$ initialized
\State $W \gets C_{1,0}(A,K)$
\Statex \hspace{1em} [AIN: $W_{i_{a} k_{a} i_{l_{1}}\ldots i_{l_{k}} i_{in} i_{f}} \gets \sum_{j} A_{i_{a} j k_{a}} K_{j i_{l_{1}}\ldots i_{l_{k}} i_{in} i_{f}}$]
\State $W \gets \tau_{p=(0\!\rightarrow\!k\!+\!2)}(W)$
\Statex \hspace{1em} [AIN: $W_{k_{a} i_{l_{1}}\ldots i_{l_{k}} i_{a} i_{in} i_{f}} \gets W_{p(i_{a} k_{a} i_{l_{1}}\ldots i_{l_{k}} i_{in} i_{f})}$]
\State $W \gets Rs_{k+2}(W)$
\Statex \hspace{1em} [AIN: $W_{k_{a} i_{l_{1}}\ldots i_{l_{k}} j i_{f}} \gets W_{k_{a} i_{l_{1}}\ldots i_{l_{k}} Rs(i_{a},i_{in}) i_{f}}$]
\State $temp \gets []$
\For {$i\in\{0,\ldots,al-1\}$}
  \State $temp \gets temp \cup \{\,convkD(X, W_{i\ldots})\,\}$
\EndFor
\State $Output_{i_{b} i_{1}\ldots i_{k} h} \gets K_{k+1}(temp)$
\If {bias is True}
  \State $Output \gets Output + b$
\EndIf
\If {activation is True}
  \State $Output \gets \sigma(Output)$
\EndIf
\end{algorithmic}
\label{Algorithm_hyperKconv}
\end{algorithm}

Figure~\ref{fig:hyperconv_indices} summarizes the index transformations and the per-component
convolution pipeline used in Algorithm~2.

\begin{figure}[htb]
\centering
\begin{tikzpicture}[
  node distance=1.4cm and 2.4cm,
  every node/.style={
    draw,
    rounded corners,
    align=center,
    text width=0.40\linewidth,
    minimum height=1.1cm,
    inner sep=4pt,
    font=\small
  },
  arrow/.style={->, thick}
]

\node (K) {Kernel tensor\\
$K_{\,i_{a}\, i_{l_1}\ldots i_{l_k}\, i_{in}\, i_f}$};

\node (A) [right=of K] {Algebra tensor\\
$A_{\,i_{a}\, j_{a}\, k_{a}}$};

\node (C) [below=of K] {Contraction $C_{1,0}$\\
$W_{\,i_{a}\, k_{a}\, i_{l_1}\ldots i_{l_k}\, i_{in}\, i_f}$};

\node (P) [right=of C] {Permutation $\tau$\\
move $k_{a}$ to front\\
$W_{\,k_{a}\, i_{l_1}\ldots i_{l_k}\, i_{a}\, i_{in}\, i_f}$};

\node (R) [below=of C] {Reshape $Rs$\\
$(i_{a},i_{in}) \mapsto j$\\
$W_{\,k_{a}\, i_{l_1}\ldots i_{l_k}\, j\, i_f}$};

\node (Loop) [right=of R] {For each $k_{a}$:\\
$Y^{(k_{a})} \gets convkD(X,\,W_{k_{a},\ldots})$};

\node (Cat) [below=of R] {Concatenation $K_{k+1}$\\
$Output \gets concat\big(\{Y^{(k_{a})}\}\big)$};

\node (Act) [right=of Cat] {Optional\\
bias + activation};

\draw[arrow] (K) -- (C);
\draw[arrow] (A) -- (C);

\draw[arrow] (C) -- (P);
\draw[arrow] (P) -- (R);

\draw[arrow] (R) -- (Loop);
\draw[arrow] (Loop) -- (Cat);
\draw[arrow] (Cat) -- (Act);

\end{tikzpicture}
\caption{Index transformations in the HyperConv layer (Algorithm~2).
The algebra multiplication tensor is contracted with the convolutional kernel,
then permuted and reshaped to obtain per-component real-valued kernels.
A standard $k$-dimensional convolution is applied independently for each algebra component,
and the results are concatenated to form the layer output.}
\label{fig:hyperconv_indices}
\end{figure}

The above formulation is implemented for $k\in\{1,2,3\}$ in the \texttt{HyperConv1D}, \texttt{HyperConv2D}, and \texttt{HyperConv3D} classes of the \textit{Hypercomplex Keras} library~\cite{HypercomplexKeras} and described in detail in~\cite{NiemczynowiczKycia}. The proposed approach extends the results of~\cite{Marcos1} from four-dimensional algebras to arbitrary finite-dimensional algebras.

The FLOP complexity is presented in the following theorem.
\begin{theorem}
The FLOP complexity for a hypercomplex-valued convolutional layer, assuming $b>a$, is
   \begin{equation}
       \mathcal{O}(b \cdot a \cdot \Pi_{l=1}^{k}L_{l} \cdot n \cdot F ),
   \end{equation}
compared with the complexity of the corresponding real-valued convolutional layer
   \begin{equation}
       \mathcal{O}(b \cdot \Pi_{l=1}^{k}L_{l} \cdot n \cdot F).
   \end{equation}
\end{theorem}
\begin{proof}
The first contraction in the Algorithm \ref{Algorithm_hyperKconv} gives $\mathcal{O}(a^{3} \cdot \Pi_{l=1}^{k}L_{l} \cdot in \cdot F ) = \mathcal{O}(a^{3} \cdot \Pi_{l=1}^{k}L_{l} \cdot n \cdot F )$ since $n=a\cdot in$.

The loop over algebra size $a$ has a complexity of a real-valued convolutional neural network, $\mathcal{O}(b\cdot \Pi_{l=1}^{k}L_{l} \cdot n \cdot F)$. Therefore this part has complexity $\mathcal{O}(a \cdot b\cdot \Pi_{l=1}^{k}L_{l} \cdot n \cdot F)$.

Taking into account that $b>a$, the complexity of the second part is dominant, which yields the result.
\end{proof}
Again, for $a=1$ both, results coincides.

\subsection{Algebraic form and expressivity of hypercomplex-valued layers}
\label{AlgHyperLayers}

The hypercomplex-valued dense and convolutional layers introduced above define a class of functions whose structure is fully determined by tensor contractions associated with algebra multiplication, followed by standard nonlinear activations. In this subsection, we make this observation explicit in order to prepare the theoretical analysis presented in the next section.

For a hypercomplex-valued dense layer, the pre-activation output corresponding to a single sample can be written in abstract index notation as
\begin{equation}
z^{k} =
\sum_{i,j,u}
A^{k}_{ij}\, x^{i}_{u}\, w^{j}_{u},
\end{equation}
where $x^{i}_{u}$ denotes the $i$-th algebra component of the input associated with multiplicity index $u$, $w^{j}_{u}$ are the corresponding learnable parameters, and $A^{k}_{ij}$ are the structure constants of the algebra. The nonlinear activation function $\sigma$ is applied component-wise after this contraction, yielding
\begin{equation}
y^{k} = \sigma(z^{k}).
\end{equation}

An analogous representation holds for hypercomplex-valued convolutional layers, where the linear part of the transformation consists of spatial convolutions applied independently to algebra components, combined through the same algebra multiplication tensor. In all cases, the hypercomplex-valued layer computes a finite sum of multilinear forms determined by the algebra structure, followed by a nonlinear mapping.

For a fixed algebra dimension $\dim(V)=al$, each hypercomplex-valued layer admits a real-valued representation acting on $\mathbb{R}^{al}$-valued inputs with a highly structured weight-sharing pattern induced by the tensor $A$. Consequently, hypercomplex-valued neural networks can be viewed as constrained real-valued networks whose expressive power is governed jointly by the choice of activation function and the algebraic structure.

This observation provides a direct link between the tensor-based formulation of hypercomplex-valued layers and classical approximation results for real-valued neural networks. In particular, it allows universal approximation properties to be studied by lifting known results to the hypercomplex-valued setting via the real representation induced by the algebra multiplication tensor. This connection is exploited in the next section, where a tensor version of the universal approximation theorem for hypercomplex-valued perceptrons is established.

Within the present framework, the theoretical analysis and the universal approximation result are restricted to component-wise (split) activation functions.

\section{Tensor version of the universal approximation theorem for a single layer of hypercomplex-valued perceptrons}\label{sec_UniversalApproximationTheorem}

In this section, we present a proof of the universal approximation theorem formulated within the tensor framework introduced above. This formulation leads to a streamlined argument that makes explicit use of the algebraic structure. The proof follows the approach presented in~\cite{BookHypercomplexNN}. For an alternative vector--matrix formulation, see~\cite{Marcos_UniversalApproximationTheorem}, which also builds on~\cite{BookHypercomplexNN}.

Throughout this section, the algebra $(V,+,\cdot)$ is assumed to be finite-dimensional and non-degenerate.

\begin{definition}
Let $(V,+,\cdot)$ be a finite-dimensional real algebra with multiplication tensor $A$ and fixed ordered basis $E=(e_0,\ldots,e_n)$. Let $\pi_i:V\to\R$ denote the corresponding coordinate projections. The algebra is (component-wise) non-degenerate with respect to $E$ if
\begin{equation}\label{eq:def}
\begin{gathered}
 \forall i\in\{0,\ldots,n\}\quad\forall x\in V:\\
 \Bigl[\bigl(\forall y\in V:\ \pi_i(A(x,y))=0\bigr)
 \lor\bigl(\forall y\in V:\ \pi_i(A(y,x))=0\bigr)\Bigr]
 \Longrightarrow x=0.
\end{gathered}
\end{equation}
Equivalently, each bilinear form $(x,y)\mapsto\pi_i(A(x,y))$ is non-degenerate. In tensor notation, this condition is equivalent to
\begin{equation}\label{eq:slices}
 B_i=\bigl(A^i{}_{pq}\bigr)_{p,q=0}^{n}
    =\bigl(\pi_i(e_p\cdot e_q)\bigr)_{p,q=0}^{n},
 \qquad \det B_i\ne 0\quad (i=0,\ldots,n).
\end{equation}
The matrix $B_i$ represents the corresponding bilinear form in the chosen basis. This is the non-degeneracy condition of Valle et al.~\cite[Definition~2]{Marcos_UniversalApproximationTheorem} expressed in tensor notation.
\end{definition}

The classical Riesz representation theorem plays a crucial role in the proof.

\begin{lemma}[Riesz representation theorem {\cite{FunctionalAnalysis}}]\label{Riesz}
Let $H$ be a Hilbert space over a field $\mathbb{F}$ (e.g., $\mathbb{R}$ or $\mathbb{C}$) with inner product $\langle \cdot \mid \cdot \rangle : H \times H \to \mathbb{F}$. Then for every continuous linear functional $\phi \in H^{*}$ there exists a unique vector $v_{\phi}\in H$ such that
\[
\phi(x) = \langle v_{\phi} \mid x \rangle
\]
for all $x \in H$.
\end{lemma}

For $H=\mathbb{R}^{N}$, endowed with the standard inner product $\langle x \mid y \rangle := x^{T}y$, every linear functional is continuous.

To adapt this result to vectors with entries in a hypercomplex algebra $V$, we regard $V^{N}$ as a real vector space. We introduce the projection onto algebra components
\[
\pi_{k}\!\left(\sum_{i=0}^{n} x^{i} e_{i}\right) = x^{k}, \qquad 0 \leq k \leq n,
\]
where $\{e_{0},\ldots,e_{n}\}$ is a fixed basis of $V$.

We now state and prove a version of Lemma~\ref{Riesz} adapted from~\cite{BookHypercomplexNN}.

\begin{lemma}[Riesz representation for non-degenerate algebras]
\label{Riesz_algebra}
Let $\phi: V^{N} \rightarrow \mathbb{R}$ be a real-valued linear functional, where $V$ is a non-degenerate algebra. Then there exist uniquely determined vectors $\{v_{i}\}_{i=0}^{n}$ with $v_{i}\in V^{N}$ such that for every $x\in V^{N}$
\[
\phi(x) = \pi_{i}\big(\langle v_{i} \mid x \rangle\big),
\qquad i=0,\ldots,n,
\]
where $\langle v \mid x \rangle = \sum_{k=1}^{N} v_{k}\cdot x_{k} = v^{T}x \in V$.
\end{lemma}

\begin{proof}
Fix $i\in\{0,\ldots,n\}$ and define a linear mapping
\[
\Phi: V^{N} \rightarrow (V^{N})^{*}, \qquad
\Phi(v_{i}) = \pi_{i} \circ \langle v_{i} \mid \cdot \rangle .
\]
Since $\dim V^{N} = \dim (V^{N})^{*}$, it suffices to show that $\Phi$ is injective.

Assume that $\Phi(v_{i})(x)=0$ for all $x\in V^{N}$. Consider the vectors
\[
y_{pq} = e_{p} E_{q}, \qquad 0\leq p \leq n,\; 1\leq q \leq N,
\]
where $E_{q}$ denotes the $q$-th canonical basis vector of $\mathbb{R}^{N}$. The family $\{y_{pq}\}$ forms a basis of $V^{N}$.

From
\[
0 = \pi_{i}(\langle v_{i} \mid y_{pq} \rangle)
\]
for all $p,q$, and the non-degeneracy of $V$, it follows that all components of $v_{i}$ vanish, hence $v_{i}=0$. Uniqueness follows immediately.
\end{proof}

We now state the universal approximation theorem for hypercomplex-valued functions.

\begin{theorem}[Universal approximation theorem]
Let $X\subset V^{N}$ be compact and let $g:X\rightarrow V$ be continuous. Then for every $\varepsilon>0$ there exist real coefficients $\alpha_{1},\ldots,\alpha_{p}$, vectors $v_{1},\ldots,v_{p}\in V^{N}$, and algebra elements $\Theta_{1},\ldots,\Theta_{p}\in V$ such that
\[
\sup_{x\in X}
\left\|
g(x) - \sum_{k=1}^{p} \alpha_{k} f(\langle v_{k}\mid x\rangle + \Theta_{k})
\right\| < \varepsilon,
\]
where the activation function $f:V\rightarrow V$ is defined component-wise by
\[
f(x)=\sum_{i=0}^{n} \sigma(\pi_{i}(x))\,e_{i},
\]
$\sigma$ is the sigmoid function, and $\|\cdot\|$ is any norm on $V$ regarded as a finite-dimensional real vector space.
\end{theorem}

\begin{proof}
Let $g:X\rightarrow V$ be continuous. It admits the decomposition
\[
g = \sum_{i=0}^{n} e_{i}\,\pi_{i}\circ g,
\]
where each component $\pi_{i}\circ g$ belongs to $C^{0}(X,\mathbb{R})$.

By the classical universal approximation theorem~\cite{Cybenko} (see also~\cite{UAP_Barron}--\cite{UAP_Leshno}), for each component $\pi_{i}\circ g$ and any $\varepsilon>0$ there exist real coefficients $a^{i}_{k}$, real-valued linear functionals $\phi^{i}_{k}$ on $V^{N}$, and real parameters $\theta^{i}_{k}$ such that
\[
\left|
\pi_{i}(g(x)) - \sum_{k=1}^{p} a^{i}_{k}\,\sigma(\phi^{i}_{k}(x)+\theta^{i}_{k})
\right| < \varepsilon
\]
for all $x\in X$.

By Lemma~\ref{Riesz_algebra}, each functional $\phi^{i}_{k}$ can be written as
\[
\phi^{i}_{k} = \pi_{i}\circ \langle v^{i}_{k}\mid \cdot \rangle
\]
for some $v^{i}_{k}\in V^{N}$. Define
\[
\Theta^{i}_{k}(\lambda)=\sum_{j\neq i} \lambda e_{j} + \theta^{i}_{k} e_{i}.
\]
Using the boundedness and monotonicity of the sigmoid function, the convergence
\[
\lim_{\lambda\to -\infty}
f(\langle v^{i}_{k}\mid x\rangle + \Theta^{i}_{k}(\lambda))
=
\sigma(\phi^{i}_{k}(x)+\theta^{i}_{k})\,e_{i}
\]
is uniform on $X$. Consequently, for sufficiently negative $\lambda$,
\[
\sup_{x\in X}
\left\|
\pi_{i}(g(x))e_{i}
-
\sum_{k=1}^{p}
\alpha_{k}\,
f(\langle v^{i}_{k}\mid x\rangle + \Theta^{i}_{k}(\lambda))
\right\| < 2\varepsilon .
\]

Summing over all components completes the proof.
\end{proof}

The above argument follows the structure of Theorem~5.5.1 in~\cite{BookHypercomplexNN} and reduces the functional-analytic aspects of the proof to the classical universal approximation theorem of~\cite{Cybenko}.

\section{Conclusion}
\label{Concl}

In this work, we presented a fully tensorial formulation of dense and convolutional neural network layers based on hypercomplex and general finite-dimensional algebras. The key observation underlying the proposed framework is the representation of algebra multiplication by a rank-three tensor. This identification allows all neural network operations to be expressed exclusively in terms of standard tensor contractions, permutations, and reshaping operations.

The resulting formulation unifies hypercomplex-valued neural network architectures across arbitrary algebra dimensions and provides a clear mathematical correspondence between algebraic structure and tensor-based computation. This perspective leads to a streamlined description of hypercomplex-valued dense and convolutional layers and enables their efficient implementation using optimized tensor primitives available in modern deep learning frameworks such as \textit{TensorFlow} and \textit{PyTorch}.

The tensorial framework presented in the paper provides a basis for future extensions to more advanced architectures, including recurrent neural networks, LSTM-based models, attention mechanisms, and Transformers.

Building on this tensor-based representation, we also established a universal approximation theorem for single-layer hypercomplex-valued perceptrons under mild non-degeneracy assumptions on the underlying algebra. The proof reduces the hypercomplex-valued approximation problem to classical real-valued results via component-wise projections, thereby providing a rigorous theoretical foundation for the proposed architectures.

The presented framework generalizes earlier constructions developed for four-dimensional algebras~\cite{Marcos1} and quaternion-valued neural networks~\cite{QuaternionicNN2} to arbitrary finite-dimensional algebras in a dimension-independent manner. This formulation is compatible with automatic optimization mechanisms available in modern libraries for efficient tensor computations. Practical implementations of the proposed models are provided in the \textit{Hypercomplex Keras} library~\cite{HypercomplexKeras} and described in detail in~\cite{NiemczynowiczKycia}, where consistency with existing quaternion-based formulations is also verified.

The universal approximation result established in this work applies to finite-dimensional non-degenerate algebras, for which algebra multiplication does not admit nontrivial null directions. In practice, this condition can be verified directly from the multiplication tensor $A_{ij}^{k}$ by checking that no nonzero element annihilates all others under left or right multiplication.

While the proposed tensor-based architectures can be implemented for more general algebraic structures, including degenerate algebras, the universal approximation guarantee does not follow from the present theorem in such cases.

\appendix

\section{Algebraic and tensorial background}
\label{app:algebra_tensor}

\subsection{Tensor product as a quotient space}
\label{app:tensor_quotient}

Let $V_{1}$ and $V_{2}$ be vector spaces over the field $\mathbb{F}$. The tensor product
$V_{1}\otimes V_{2}$ is defined as the quotient space
\[
V_1 \otimes V_2 := F(V_1 \times V_2)/L,
\]
where $F(V_1 \times V_2)$ denotes the free vector space over $\mathbb{F}$ generated by the Cartesian product $V_1 \times V_2$ (taking the quotient of $V_1 \times V_2$ directly as a vector space would yield the zero space), and where $L\subset V_{1}\times V_{2}$ is the subspace spanned by the elements
\[
(v+w,x)-(v,x)-(w,x),\quad
(v,x+y)-(v,x)-(v,y),
\]
\[
(\lambda v,x)-\lambda(v,x),\quad
(v,\lambda x)-\lambda(v,x),
\]
for all $v,w\in V_{1}$, $x,y\in V_{2},$ and $\lambda\in\mathbb{F}$.
The equivalence class of $(v,x)$ is denoted by $v\otimes x$.


\subsection{Universal factorization theorem}
\label{app:universal_factorization}

Let $F:V\times W\rightarrow\mathbb{F}$ be a bilinear mapping. Then there exists a unique
linear mapping $f:V\otimes W\rightarrow\mathbb{F}$ such that
\[
F = f\circ \otimes .
\]

The relationship is summarized in the commutative diagram

\[
\xymatrix{
V\times W \ar[dr]^{F} \ar[rr]^{\otimes} & & V\otimes W \ar@{-->}[dl]^{f} \\
& \mathbb{F} &
}
\]

The theorem extends directly to multilinear mappings.


\subsection{Extended examples of tensors}
\label{app:tensor_examples}

\paragraph{Bilinear mappings.}
Let $F:V\times W\rightarrow\mathbb{F}$ be a bilinear mapping and let $\{e_{i}\}$ and $\{f_{j}\}$
be bases of $V$ and $W$, respectively. Then
\[
F(e_{i},f_{j}) = F_{ij},
\qquad
f = F_{ij}\, e^{i}\otimes f^{j}.
\]
For $u=u^{i}e_{i}$ and $x=x^{j}f_{j}$,
\[
F(u,x)=F_{ij}u^{i}x^{j}.
\]

\paragraph{Linear operators.}
For a linear operator $T:V\rightarrow W$, there exists a canonical identification
\[
L(V,W)\simeq V^{*}\otimes W.
\]
In a fixed basis,
\[
T = T^{j}_{i}\, e^{i}\otimes f_{j}.
\]


\subsection{Functoriality of the tensor product}
\label{app:functoriality}

Let $f:V\rightarrow W$ be a linear map. The induced mapping
\[
\otimes^{n}f:\otimes^{n}V\rightarrow\otimes^{n}W
\]
is defined by
\[
\otimes^{n}f(v_{1}\otimes\cdots\otimes v_{n})
=
f(v_{1})\otimes\cdots\otimes f(v_{n}),
\]
and extended by linearity.


\subsection{Formal definition of broadcasting}
\label{app:broadcasting}

Let $A:V\rightarrow W$ be a linear operator. Broadcasting is defined as
\[
b_{1}^{n}: L(V,W)\rightarrow L(V^{n},W^{n}),
\]
where
\[
b_{1}^{n}(A)(v_{1},\ldots,v_{n})=(Av_{1},\ldots,Av_{n}).
\]

\section{Computational remarks}
\label{app:computational}

Hypercomplex-valued neural network layers are constructed exclusively from standard tensor
operations, including tensor contractions, index permutations, reshaping, and classical
real-valued matrix multiplication or convolution.

The dominant computational cost arises from contractions involving the algebra
multiplication tensor, followed by standard dense or convolutional operations. For fixed
algebra dimension $\dim(V)$, the computational complexity remains comparable to that of
real-valued neural networks operating on structured multi-channel representations.

All operations admit fully vectorized implementations and are compatible with optimized
GPU and TPU backends. The computational overhead introduced by hypercomplex-valued layers is
explicit and controlled by the algebra dimension.

\paragraph{Scaling with the algebra dimension}
For the \texttt{HyperDense} layer, the dominant computational cost arises from the contraction
between the algebra multiplication tensor $A_{ij}^{k}$ and the kernel tensor $K$,
followed by a standard matrix multiplication after reshaping.
For a fixed batch size, this cost scales linearly with the number of units and
quadratically with the algebra dimension $\dim(V)$, reflecting the bilinear nature
of algebra multiplication encoded in $A$.

For \texttt{HyperConv} layers, the cost per output filter scales proportionally to the
corresponding real-valued convolution multiplied by a factor that depends on the algebra
dimension. As shown in Algorithm~2, the convolution is applied independently to each
algebra component, resulting in a computational cost proportional to
$\dim(V)$ times the cost of the underlying real-valued convolution with the same kernel
size and number of filters.

In both cases, the scaling with $\dim(V)$ is explicit and does not introduce any hidden
computational overhead beyond that induced by the algebra multiplication tensor.

\section*{Supplementary information}

\section*{Declarations}

\begin{itemize}
\item Funding - This paper has been supported by the Polish National Agency for Academic Exchange Strategic Partnership Programme under Grant No. BPI/PST/2021/1/00031 (nawa.gov.pl).

\item Conflict of interest/Competing interests - Not applicable
\item Ethics approval and consent to participate - Not applicable
\item Consent for publication - All authors agree on publication
\item Data availability - Not applicable
\item Materials availability - Not applicable
\item Code availability - Not applicable
\item Author contribution - 
A.N., R.K.: Conceptualization, Methodology, Investigation, Writing -
Original Draft, Review, Editing. A.N.: Funding acquisition, Supervision.

\end{itemize}

\section*{Acknowledgements}
This paper has been supported by the Polish National Agency for Academic Exchange Strategic Partnership Programme under Grant No. BPI/PST/2021/1/00031 (nawa.gov.pl).

RK and AN are acknowledge the support of COST CA24122 action.

We would like to thanks prof. Jin Hee Yoon from Sejong University for hospitality during our stay in South Korea.


\end{document}